\documentclass[final]{cvpr}

\usepackage{times}
\usepackage{epsfig}
\usepackage{graphicx}
\usepackage{amsmath}
\usepackage{amssymb}
\usepackage[utf8]{inputenc}
\usepackage[T1]{fontenc}
\usepackage{url}
\usepackage{booktabs}
\usepackage{amsfonts}
\usepackage{nicefrac}
\usepackage{microtype}
\usepackage{xcolor}
\usepackage{tabularx}
\usepackage{graphicx}
\usepackage{amsmath}
\usepackage{subcaption}
\usepackage{tabulary,multirow,overpic,xcolor}
\definecolor{mygray}{gray}{0.92}
\definecolor{baselinecolor}{gray}{.9}

\usepackage{listings}
\usepackage{amsthm}
\usepackage{tabulary}
\usepackage{colortbl}
\usepackage{enumitem}
\usepackage{braket}
\usepackage{array}
\usepackage{float}
\usepackage{enumitem}
\usepackage{booktabs}
\usepackage{algorithm}
\usepackage{bbding}
\usepackage{listings}
\usepackage{tabu}
\usepackage{booktabs}
\usepackage{multirow}
\usepackage{tabularx}
\usepackage{enumitem}
\usepackage{graphicx}
\usepackage{caption}
\usepackage{wrapfig}
\usepackage{lipsum}

\usepackage[pagebackref,breaklinks,colorlinks]{hyperref}

\def\cvprPaperID{****}
\def\confName{CVPR}
\def\confYear{2021}
\graphicspath{{./figs/}{./figs/result/}}

\usepackage{bbold}
\usepackage{multirow}
\usepackage{bm}
\usepackage{arydshln}
\usepackage{xspace}
\usepackage{url}

\makeatletter
\newcommand{\figcaption}[1]{\def\@captype{figure}\caption{#1}}
\newcommand{\tblcaption}[1]{\def\@captype{table}\caption{#1}}
\newcommand{\customparagraph}[1]{\par{\noindent\textbf{#1:}}}
\newcommand{\textcite}[1]{``\textit{#1}''}

\makeatother

\def\chi{Proceedings of the SIGCHI Conference on Human Factors in Computing Systems (CHI)}

\makeatletter
\DeclareRobustCommand\onedot{\futurelet\@let@token\@onedot}
\def\@onedot{\ifx\@let@token.\else.\null\fi\xspace}
\def\eg{\emph{e.g}\onedot} 
\def\ie{\emph{i.e}\onedot}

\def\etal{\emph{et al}\onedot}

\makeatother
\begin{document}

\title{Pre-Training for 3D Hand Pose Estimation with Contrastive Learning on Large-Scale Hand Images in the Wild}

\author{
Nie Lin$^*$, Takehiko Ohkawa$^*$, 
Mingfang Zhang, \\ Yifei Huang, Ryosuke Furuta, Yoichi Sato\\
Institute of Industrial Science\\
The University of Tokyo\\
{\tt\small \{nielin, ohkawa-t, mfzhang, hyf, furuta, ysato\}@iis.u-tokyo.ac.jp}}

\newcommand\blfootnote[1]{%
  \begingroup
  \renewcommand\thefootnote{}\footnote{#1}%
  \addtocounter{footnote}{-1}%
  \endgroup
}

\maketitle

\blfootnote{
$^*$Equal contribution
}
\begin{abstract}
We present a contrastive learning framework based on in-the-wild hand images tailored for pre-training 3D hand pose estimators, dubbed HandCLR. Pre-training on large-scale images achieves promising results in various tasks, but prior 3D hand pose pre-training methods have not fully utilized the potential of diverse hand images accessible from in-the-wild videos. To facilitate scalable pre-training, we first prepare an extensive pool of hand images from in-the-wild videos and design our method with contrastive learning. Specifically, we collected over 2.0M hand images from recent human-centric videos, such as \textit{100DOH} and \textit{Ego4D}. To extract discriminative information from these images, we focus on the \textit{similarity} of hands; pairs of similar hand poses originating from different samples, and propose a novel contrastive learning method that embeds similar hand pairs closer in the latent space. Our experiments demonstrate that our method outperforms conventional contrastive learning approaches that produce positive pairs sorely from a single image with data augmentation. We achieve significant improvements over the state-of-the-art method in various datasets, with gains of 15\% on FreiHand, 10\% on DexYCB, and 4\% on AssemblyHands.
\end{abstract}
\section{Introduction}
Hands are a trigger for us to interact with the world, as seen in various human-centric videos. Precise recognition of hand states, such as 3D keypoints, is crucial for video understanding~\cite{sener:cvpr22,wen:arxiv23}, AR/VR interfaces~\cite{han:tog22,wu:vcir20}, and robot learning~\cite{chao:cvpr21,qin:eccv22}. To this end, 3D hand pose estimation has been studied through constructing labeled datasets~\cite{ohkawa:ijcv23,zimmermann:iccv19,chao:cvpr21,ohkawa:cvpr23} and advancing pose estimators~\cite{cai:eccv18,ge:cvpr19,park:cvpr22,liu:cvpr24,fan:eccv24}. However, utilizing large-scale, unannotated hand videos for pre-training remains underexplored, while vast collections of such videos, like 3,670 hours of videos from Ego4D~\cite{grauman:cvpr22} and 131 days from 100DOH~\cite{shan:cvpr20}, are available.

Some works utilize unlabeled hand images for 3D hand pose pre-training using contrastive learning like SimCLR~\cite{chen:icml20}, which maximizes agreement between positive pairs while repelling negatives. Spurr~\etal~\cite{spurr:iccv21} introduce pose equivariant contrastive learning (PeCLR) by aligning geometry in latent space after affine transformations for input images. However, both SimCLR and PeCLR create positive pairs from a single sample by applying augmentation, limiting the gains from positive pairs as their hand appearance and backgrounds are identical. Ziani~\etal~\cite{ziani:3dv22} extend the contrastive learning framework to video sequences by treating temporally adjacent frames as positive pairs. However, in-the-wild videos can challenge tracking hands across frames, especially in egocentric views where hands may be unobservable due to camera motion.
In addition, adjacent frames still pose limited appearance variation of hands and backgrounds.
\begin{figure*}[t!]
    \begin{center}
    \includegraphics[width=1.00\textwidth]{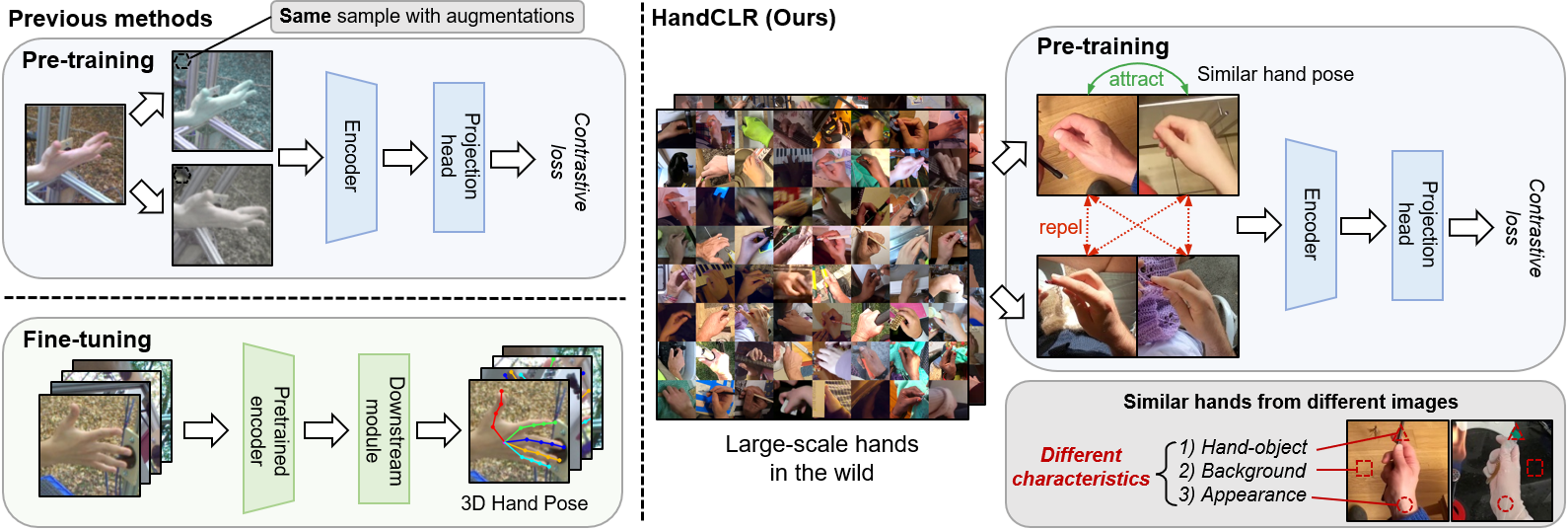}
    \end{center}
   \caption{
   \textbf{The pipeline of pre-training and fine-tuning in 3D hand pose estimation.} \textbf{(Left)} Previous pre-training methods (\eg, PeCLR~\cite{spurr:iccv21}) learn from positive pairs originating from the same with different augmentations and fine-tune the network on a dataset.
    \textbf{(Right)} Our method is designed to learn from positive pairs with similar foreground hands, sampled from a pool of hand images in the wild. 
   }
\label{fig:pipline}
\end{figure*}

In this work, we introduce a novel contrastive learning framework for 3D hand pose pre-training to leverage diverse hand images in the wild, with the largest 3D hand pose pre-training set to date. We collected 2.0M hand images from in-the-wild videos, specifically from Ego4D~\cite{grauman:cvpr22} and 100DOH~\cite{shan:cvpr20}, using an off-the-shelf hand detector~\cite{lugaresi:arxiv19}. Our pre-training set significantly exceeds the scale of prior works by two orders of magnitude, such as the 32-47K images in~\cite{spurr:iccv21} and 86K images from 100DOH in~\cite{ziani:3dv22}.

Our method focuses on learning discriminative information by leveraging the similarity of hands from different domains. Unlike SimCLR and PeCLR, we observe that it is further informative to learn from positive pairs with similar foreground hands but from different images. As shown in Fig.~\ref{fig:pipline}, our positive pairs based on different images offer additional information gains from different types of object interactions, backgrounds, and hand appearances. Specifically, we use an off-the-shelf 2D hand pose estimator~\cite{lugaresi:arxiv19} to identify similar hands from the pre-training set.

Our contributions are threefold: 1) We construct a large-scale in-the-wild hand dataset for 3D hand pose pre-training. 2) We propose a pre-training framework using similar hand pairs via contrastive learning. 3) Our model achieves state-of-the-art performance across multiple datasets.
\vspace{-3pt}
\section{Related Work}
\customparagraph{3D hand pose estimation} The task of 3D hand pose estimation aims to regress 3D keypoints of hand joints. Annotating 3D hand poses is challenging, which allows us to have limited labeled datasets~\cite{ohkawa:ijcv23}, mostly constructed in controlled laboratory settings~\cite{zimmermann:iccv19,chao:cvpr21,moon:eccv20,ohkawa:cvpr23}. Given this challenge, two approaches have been proposed to facilitate learning from limited annotations: pseudo-labeling~\cite{liu:cvpr21,ohkawa:eccv22,yang:iccv21,liu:cvpr24} and self-supervised pre-training~\cite{spurr:iccv21,ziani:3dv22}. Pseudo-labeling methods learn from pseudo-ground-truth assigned on unlabeled images~\cite{liu:cvpr21,yang:iccv21,ohkawa:eccv22,liu:cvpr24}. Alternatively, pre-training methods first pre-train an encoder with contrastive learning on unlabeled images and then fine-tune on labeled images~\cite{spurr:iccv21,ziani:3dv22}. While prior works use relatively small pre-training sets (\eg, 32-47K images in \cite{spurr:iccv21} and 86K images in \cite{ziani:3dv22}), our work emphasizes leveraging in-the-wild images on a large scale. We collected hand images from large human-centric datasets such as Ego4D~\cite{grauman:cvpr22} and 100DOH~\cite{shan:cvpr20}, expanding our pre-training set to 2.0M images.

\customparagraph{Contrastive learning} 
The framework of contrastive learning has emerged as a powerful self-supervised learning, bringing positive samples closer while pushing negative samples apart~\cite{chopra:cvpr05, schroff:cvpr15, ohsong:cvpr16, sohn:nips16, he:arxiv18, he:cvpr20}. Standard methods generate positive samples from individual images with data augmentation (\ie, self-positives)~\cite{grill:neurips20, caron:neurips20, chen_2:cvpr21, radford:icml21, caron:iccv21}, but restrict the learning of explicit relationships between samples. To address this, Zhang~\etal propose a relaxed extension of self-positives, \textit{non-self-positives}~\cite{zhang:eccv22}, which share similar characteristics (\eg, same scene~\cite{Arandjelovic:cvpr16, ge:eccv20, berton:cvpr22, Hausler:cvpr21} or instance~\cite{chen:iccv21, chen_3:cvpr21, ge:neurips20, gu:cvpr23}) but originate different images. This enables the incorporation of diverse inter-sample consistency and facilitates the learning of semantics more easily. Skeleton-based action recognition methods identify non-self-positives by searching similar human skeletons~\cite{zhang:eccv22}, whereas it relies on online mining, increasing computational overhead in training. In contrast, our approach creates non-self-positives using 2D hand keypoints offline for avoiding the overhead and scaling pre-training with any large data.
\vspace{-2pt}
\section{Method}
Our approach called HandCLR aims to pre-train an encoder of a 3D hand pose estimator with large human-centric videos available in the wild. We first construct the pre-training set from egocentric and exocentric hand videos (Sec.~\ref{sec:preproc}), then find similar hand images to define positive pairs (Sec.~\ref{sec:simhand}), and finally incorporate these positive pairs into a contrastive learning framework (Sec.~\ref{sec:contras}).

\subsection{Data preprocessing}\label{sec:preproc}
Our preprocessing involves creating a set of valid hand images for pre-training, which is sampled from a dataset with $N$ videos, $\{ v_1, v_2, \dots, v_N\}$. We use an off-the-shelf hand detector~\cite{shan:cvpr20} to select valid frames with hands. Given an image of a video, $I_{\textrm{full}}\in v_i$, the model detects the existence of the hand and gives hand crops enclosing either hand identity (right/left) from $I_{\textrm{full}}$. To avoid bias regarding hand identity, we balance the number of right and left hand crops and convert all crops to the right, allowing us to address all crops equally. Then, we create a frame set for each video as \( \mathcal{F}_i = \{ I_{i,1}, I_{i,2}, \dots, I_{i,T_{i}} \} \), where \( I_{i,j} \in \mathbb{R}^{H \times W \times 3} \) represents the processed crop with height \( H \) and width \( W \), and $T_i$ is the number of crops in the video $v_i$. The height \( H \) and width \( W \) are defined post-resize to give the uniform image size. Using this frame set $\mathcal{F}_i$, the video dataset can be re-represented as $\mathcal{V} = \{ \mathcal{F}_1, \mathcal{F}_2, \dots, \mathcal{F}_N\}$. Specifically, we processed two datasets, Ego4D~\cite{grauman:cvpr22} and 100DOH~\cite{shan:cvpr20}, comprising 8K and 21K videos, respectively.

\subsection{Mining similar hands}\label{sec:simhand}
Our preliminary experiments indicate that learning from positive pairs with similar foreground hands but from different images could provide additional information gains compared to conventional contrastive learning~\cite{chen:icml20,spurr:iccv21}. Here we construct a mining algorithm for similar hands from $\mathcal{V}$ by focusing on pose similarity between hand images. We first extract 2D pose from $I$, embedding in the latent space, and design a scheme for effective positive sample mining.

\customparagraph{Pose embedding}
To compute the hand pose similarity robustly, we obtain a $D$-dimensional embedding of 2D hand keypoints, $\mathbf{p} \in \mathbb{R}^{D}$, for each image $I$. Using an off-the-shelf 2D hand pose estimator \(\phi\)~\cite{lugaresi:arxiv19}, we predict 2D keypoints for 21 joints. We use a concatenated \(42\)-dimensional vector as the output of \(\phi\) for later use. As these 2D keypoints are prone to be noisy, we apply PCA-based dimensionality reduction to project the vector into a lower-dimensional space of size \(D\). Given the PCA projection matrix \(M \in \mathbb{R}^{42 \times D}\), the pose embedding \(\mathbf{p}\) is calculated as \(\mathbf{p} = M^{T} \mathbf{\phi}(I)\). This process mitigates noise and provides a more robust representation. We empirically choose $D=14$ for our experiments.

\customparagraph{Mining}
This step is designed to identify a positive sample $J \in \mathbb{R}^{H \times W \times 3}$ paired with a query image $I$.
We denote the similarity mining logic as $J = \mathrm{SiM} (I)$. When using the closest sample in the PCA space, we encounter a trivial solution $I,J \in v_i$, where both images originate from the same video $v_i$. Similarly to~\cite{ziani:3dv22}, the supervision by positive samples from the same video have less diversity in backgrounds, hand appearances, and object interactions. Thus we are motivated to find similar hands derived from different video domains. Specifically, we search the minimum distance within the set of all frames except for $v_i$, written as $\mathcal{F}_{\mathrm{excl},i}=\bigcup_{\substack{k=1 \\ k \neq i}}^N \bigcup_{j=1}^{T_k} I_{k j}$. Given an query $I_{i,j}$ where a $j$-th image of a $i$-th video, the function $\mathrm{SiM}(\cdot)$ is formulated as
\begin{equation}
    \mathrm{SiM} (I_{i,j}) = {\arg \min} _{x \in \mathcal{F}_{\mathrm{excl},i}} d(M^{T} \phi(x), M^{T} \phi(I_{i,j})),
\end{equation}
where $d(\cdot,\cdot)$ is the Euclidean distance metric.

\begin{table*}[t]
\centering
\scalebox{0.87}{
\begin{tabular}{l|c|cc|cc|cc}
\toprule
\multirow{2}{*}{\textbf{Method}} & \multirow{2}{*}{\textbf{Pre-training}} & \multicolumn{2}{c|}{\textbf{FreiHand (Exo)}~\cite{zimmermann:iccv19}} & \multicolumn{2}{c|}{\textbf{DexYCB (Exo)}~\cite{chao:cvpr21}} & \multicolumn{2}{c}{\textbf{AssemblyHands (Ego)}~\cite{ohkawa:cvpr23}} \\ 
\cmidrule(lr){3-4} \cmidrule(lr){5-6} \cmidrule(lr){7-8}
& & \textbf{MPJPE↓} & \textbf{PCK-AUC↑} & \textbf{MPJPE↓} & \textbf{PCK-AUC↑} & \textbf{MPJPE↓} & \textbf{PCK-AUC↑}\\ 
\midrule
\multirow{2}{*}{SimCLR~\cite{chen:icml20}} 
& 100DOH-1M  & 19.30 & 85.36 & 20.13 & 83.75 & 20.01 & 84.21 \\
& Ego4D-1M & 19.36 & 85.09 & 20.22 & 83.50 & 20.32 & 83.85 \\
\midrule
\multirow{2}{*}{PeCLR~\cite{spurr:iccv21}} 
& 100DOH-1M & 19.58 & 84.71 & 18.39 & 18.39 & 19.12 & 85.64 \\
& Ego4D-1M & 19.07 & 85.62 & 18.99 & 85.40 & 19.20 & 85.57 \\
\midrule
\multirow{2}{*}{\begin{tabular}[c]{@{}c@{}}HandCLR\\ 
\end{tabular}} 
& 100DOH-1M & 16.73 & 88.66 & 17.34 & 87.84 & 18.50 & 86.56 \\ 
& Ego4D-1M & 16.15 & 89.48 & 16.99 & 88.34 & 18.26 & 86.95 \\
& Ego4D-1M+100DOH-1M & \textbf{15.79} & \textbf{90.04} & \textbf{16.71} & \textbf{88.86} & \textbf{18.23} & \textbf{86.90}   \\
\bottomrule
\end{tabular}
}
\vspace{5pt}
\caption{\textbf{Comparison with the state of the art.} We show 3D hand pose estimation accuracy (MPJPE↓) on the FreiHand (Exo)~\cite{zimmermann:iccv19}, DexYCB (Exo)~\cite{chao:cvpr21} and AssemblyHands (Ego)~\cite{ohkawa:cvpr23}. 
Our method achieves the best results across various pre-training datasets.
}
\label{tab:main_SOTA}
\end{table*}

\subsection{Contrastive learning}\label{sec:contras}
Given the positive samples $(I,J)$ constructed by Sec.~\ref{sec:simhand}, we describe feature extraction process and contrastive learning loss. Following~\cite{chen:icml20, spurr:iccv21}, we treat all samples other than its corresponding positive sample as negative samples. In our framework, feature extraction is performed by two learnable components: an encoding head \( E(\cdot) \) and a projection head \( g(\cdot) \). We define image augmentation as \( \mathbf {T} \) and the entire model as \( f = g \circ E \).
Given the positive pair \((I, J)\), feature extraction is performed as \( \mathbf{z} = f(\mathbf{T}(I)) \) and  \( \mathbf{z}^+ = f(\mathbf{T}(J)) \). Since \( \mathbf {T} \) introduces geometric transformations that may cause the misalignment between the image and the feature spaces, we correct such error with the inverse transformation \( \mathbf {T}^{-1} \) as~\cite{spurr:iccv21}. Finally, we use the \textit{NT-Xent} loss~\cite{chen:icml20} for contrastive learning, enabling the feature alignment between $\mathbf{z}$ and $\mathbf{z}^+$. For fine-tuning, we initialize our model with the pre-trained encoder $E(\cdot)$ and then fine-tune with a 3D pose regressor on labeled datasets. The 3D regressor involves 2D heatmap regression and 3D localization, inspired by DetNet~\cite{zhou:cvpr20}.
\vspace{-2pt}
\section{Experiments}
In this section, we begin by detailing the datasets and present our key experiments by comparing our results with state-of-the-art methods.

\subsection{Datasets}
\customparagraph{Pre-training datasets}
We collected a large collection of hand images from two major video datasets, Ego4D~\cite{grauman:cvpr22} and 100DOH~\cite{shan:cvpr20}, capturing egocentric and exocentric views respectively. From Ego4D, a vast egocentric video dataset with 3,670 hours of footage, we extracted 1.0M hand images from 8K videos. Similarly, from the exocentric dataset 100DOH, which includes 131 days of YouTube footage and 100K annotated hand-object interaction frames, we extracted 1.0M hand images from 20K videos. These extensive datasets provide diverse hand-object interactions across different views.

\customparagraph{Fine-tuning datasets} We conduct fine-tuning experiments on three datasets with ground-truth 3D hand pose: FreiHand~\cite{zimmermann:iccv19}, DexYCB~\cite{chao:cvpr21}, and AssemblyHands~\cite{ohkawa:cvpr23}. FreiHand, with 130K training frames, includes both green screen and real-world backgrounds, while DexYCB offers 582K images of natural hand-object interactions in a controlled laboratory setting. AssemblyHands, the largest of the three, consists of 412K training and 62K test samples, collected from egocentric perspectives in object assembly scenarios. 

\subsection{Results}
As shown in Tab.~\ref{tab:main_SOTA}, we compare our method with state-of-the-art pre-training methods for 3D hand pose estimation using the metrics of MPJPE (↓) and PCK-AUC (↑).
\customparagraph{Pre-training results}
We observe that our method significantly outperforms SimCLR and PeCLR across various datasets under the same pre-training data setup. Specifically, on the FreiHand dataset, our approach achieves a 15.3\% improvement with Ego4D-1M pre-training. Furthermore, our method demonstrates strong performance on larger datasets, with a 10.53\% gain on DexYCB and a 4.90\% improvement on AssemblyHands compared to PeCLR. These results confirm that our model consistently achieves superior performance across various pre-training datasets.

\vspace{-2pt}
\begin{figure}[t!]
\centering
\begin{minipage}[t]{0.50\textwidth} 
\centering
\footnotesize 
\begin{tabular}{cccc}
    \toprule
    \textbf{Pre-training size} & \textbf{Method} & \textbf{MPJPE ↓} & \textbf{PCK-AUC ↑} \\ 
    \midrule
    \multirow{3}{*}{Ego4D-50K} & SimCLR & 53.94 & 42.54 \\
                         & PeCLR  & 47.42 & 49.85 \\
                         & HandCLR & \textbf{35.32} & \textbf{63.35} \\ 
    \midrule
    \multirow{3}{*}{Ego4D-100K} & SimCLR & 53.49 & 43.12 \\
                          & PeCLR  & 46.00 & 51.50 \\
                          & HandCLR & \textbf{31.06} & \textbf{68.66} \\ 
    \midrule
    \multirow{3}{*}{Ego4D-500K} & SimCLR & 49.91 & 47.61 \\
                          & PeCLR  & 43.18 & 54.15 \\
                          & HandCLR & \textbf{28.27} & \textbf{72.97} \\ 
    \midrule
    \multirow{3}{*}{Ego4D-1M}   & SimCLR & 46.17 & 50.62 \\
                          & PeCLR  & 34.42 & 64.93 \\
                          & HandCLR & \textbf{23.68} & \textbf{79.62} \\ 
    \bottomrule
\end{tabular}
\vspace{5pt}
\captionof{table}{\textbf{Comparison with different pre-training data sizes.} \\
We use 10\% of the labeled FreiHand~\cite{zimmermann:iccv19} dataset for fine-tuning.
}
\label{tab:evaluation_metrics}
\end{minipage}%
\hfill
\vspace{-12pt}
\end{figure}

\customparagraph{Performance on Ego \& Exo hands}
We evaluate how pre-training with egocentric views (Ego4D) and exocentric views (100DOH) affects the performance in datasets with their corresponding views, namely AssemblyHands for egocentric and FreiHand and DexYCB for exocentric views. Interestingly, matching pre-training viewpoints does not consistently enhance performance, indicating that the view gaps have limited effects. Instead, factors like dataset diversity and the characteristics of pre-training methods are more crucial in determining effectiveness. We also assess pre-training performance using both perspectives, Ego4D and 100DOH. Combining the two datasets, the last row of Tab.~\ref{tab:main_SOTA}, leads to the best performance in all three datasets, underscoring the potential of enriching data diversity with different camera characteristics.

\customparagraph{Effect of different pre-training data sizes} 
We study results with different sizes of pre-training data, namely 50K, 100K, 500K, and 1M in Tab.~\ref{tab:evaluation_metrics}.
We specifically test the pre-trained networks on limited labeled data, \ie, 10\% of FreiHand.
This shows that HandCLR consistently improves in all settings, with gains increasing further with more pre-training data.

\begin{figure}[t!]
\centering
\begin{minipage}[t]{0.45\textwidth} 
\centering
\includegraphics[width=0.95\textwidth]{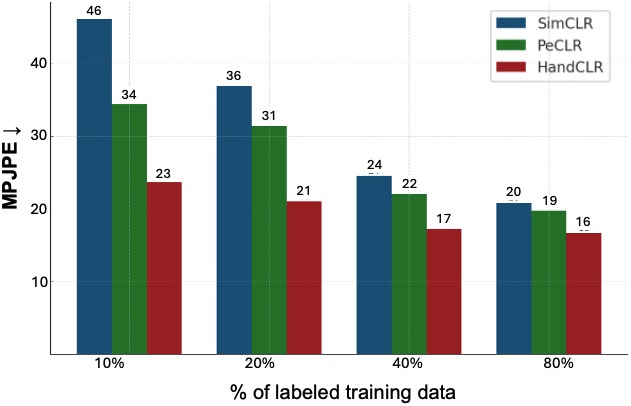}
\caption{\textbf{Comparison with different data availability in fine- \\ 
tuning.} Variations in the percentage of labeled data correspond \\ 
to different subsets of the FreiHand~\cite{zimmermann:iccv19} dataset, following the \\
experimental design in \cite{spurr:iccv21}.}
\label{fig:performance_histogram}
\end{minipage}
\hfill
\vspace{-12pt}
\end{figure}

\customparagraph{Results in smaller fine-tuning sets}
Fig.~\ref{fig:performance_histogram} illustrates the MPJPE performance comparison of three methods under different proportions of labeled data, namely 10\%, 20\%, 40\%, and 80\% in FreiHand.
The results show that our HandCLR method performs particularly well in a limited data regime, such as 10\% and 20\%, compared to the baselines.

\vspace{-3pt}
\section{Conclusion}
We introduce a contrastive learning framework for pre-training 3D hand pose estimators using the largest in-the-wild pre-training set. Our approach leverages similar hand pairs from diverse videos, significantly enhancing the information gained during pre-training over existing methods. Experiments show our method achieves state-of-the-art performance in 3D hand pose estimation across multiple datasets. This work demonstrates the benefits of pre-training with large-scale in-the-wild images and lays the foundation for future research on using diverse human-centric videos to improve the robustness of 3D hand pose estimation.

\customparagraph{Acknowledgments}
\noindent We thank Minjie Cai for helpful discussions on this manuscript. 
This work was supported by the
JST ACT-X Grant Number JPMJAX2007,
JST ASPIRE Grant Number JPMJAP2303, and
JSPS KAKENHI Grant Number 24K02956.

{\small
\bibliographystyle{ieee_fullname}
\bibliography{./refs/ref_base.bib,./refs/ref_hands.bib,./refs/ref.bib,./refs/ref_data.bib,./refs/ref_noise_label.bib,./refs/ref_cl.bib}
}
\end{document}